\documentclass[final,3p,times,authoryear]{elsarticle}

\usepackage{float}
\usepackage{amssymb}

\usepackage{amsmath}
\usepackage{multirow}
\usepackage{amsmath}
\usepackage{algorithm}
\usepackage{algorithmic}
\usepackage{arydshln}
\usepackage{float}
\journal{Engineering Applications of Artificial Intelligence}

\begin{document}

\begin{frontmatter}

\title{Safe Path Planning and Observation Quality Enhancement Strategy for Unmanned Aerial Vehicles in Water Quality Monitoring Tasks}
\author[1]{Yuanshuang Fu} %% Author name
\ead{202421230104@std.uestc.edu.cn}
\author[2]{Qianyao Wang}
\ead{marlowe@mail.ncut.edu.cn}
\author[2]{Qihao Wang}
\ead{wangqihao.ncut@gmail.com}
\author[1]{Bonan Zhang}
\ead{202421230105@std.uestc.edu.cn}
\author[2]{Jiaxin Zhao}
\ead{23101150106@mail.ncut.edu.cn}
\author[2]{Yiming Cao}
\ead{23101150119@mail.ncut.edu.cn}

\author[2]{Zhijun Li\corref{cor1}}
\ead{marlowe@mail.ncut.edu.cn}
\cortext[cor1]{Corresponding author}

\affiliation[1]{
	country={University of Electronic Science and Technology of China}}
\affiliation[2]{    country={North China University of Technology}}

%% Abstract
\begin{abstract}

Unmanned Aerial Vehicle (UAV) spectral remote sensing technology is widely used in water quality monitoring. However, in dynamic environments, varying illumination conditions, such as shadows and specular reflection (sun glint), can cause severe spectral distortion, thereby reducing data availability. To maximize the acquisition of high-quality data while ensuring flight safety, this paper proposes an active path planning method for dynamic light and shadow disturbance avoidance. First, a dynamic prediction model is constructed to transform the time-varying light and shadow disturbance areas into three-dimensional virtual obstacles. Second, an improved Interfered Fluid Dynamical System (IFDS) algorithm is introduced, which generates a smooth initial obstacle avoidance path by building a repulsive force field. Subsequently, a Model Predictive Control (MPC) framework is employed for rolling-horizon path optimization to handle flight dynamics constraints and achieve real-time trajectory tracking. Furthermore, a Dynamic Flight Altitude Adjustment (DFAA) mechanism is designed to actively reduce the flight altitude when the observable area is narrow, thereby enhancing spatial resolution. Simulation results show that, compared with traditional PID and single obstacle avoidance algorithms, the proposed method achieves an obstacle avoidance success rate of 98\% in densely disturbed scenarios, significantly improves path smoothness, and increases the volume of effective observation data by approximately 27\%. This research provides an effective engineering solution for precise UAV water quality monitoring in complex illumination environments.

\end{abstract}

%% Keywords

\end{frontmatter}


\section{Introduction}
Water resources are a fundamental element for maintaining ecological system stability and human societal sustainable development. Nevertheless, with the improper discharge of industrial wastewater, non-point agricultural pollution, and domestic sewage, global water pollution is worsening, and the risk of water quality deterioration is continuously increasing \citet{ref:1,ref:2,ref:3,ref:4}. Efficient monitoring of key water quality parameters, such as chlorophyll concentration, turbidity, and typical pollutant content, is a prerequisite for pollution tracing, risk early warning, and ecological restoration, as well as crucial support for scientific water resource management \citet{ref:5,ref:6,ref:7}. In recent years, with the continuous advancement of sensor technology and UAV platforms, UAV-based water quality monitoring methods equipped with hyperspectral or multispectral sensors have become a research hotspot \citet{ref:8,ref:9,ref:10}. Compared with traditional manual sampling or fixed monitoring stations, this method offers advantages such as high spatial resolution, wide coverage, and strong maneuverability, capable of obtaining high-precision monitoring information for large-scale water bodies in a short time \citet{ref:11,ref:12,ref:13}. Furthermore, UAV spectral monitoring features non-contact sampling, avoiding secondary disturbance to the water body, and enables the simultaneous inversion of multiple water quality indicators, including chlorophyll, suspended matter, and chemical oxygen demand, based on multi-band spectral reflectance characteristics \citet{ref:14,ref:15,ref:16}. Consequently, UAV spectral monitoring has shown unique advantages in tracking pollution processes, assessing eutrophication in lakes and reservoirs, and verifying the effectiveness of watershed ecological restoration, establishing itself as a vital component of the water quality monitoring technology system \citet{ref:17,ref:18,ref:19}.

Due to the rapid development of UAV spectral monitoring in recent years, an increasing number of studies have focused on using remote sensing imagery combined with traditional algorithms and machine learning methods to perform high-precision quantitative inversion of key water quality parameters. For instance, \citet{ref:20} mainly focused on using remote sensing imagery combined with traditional algorithms and artificial intelligence methods for large-scale quantitative inversion of various water quality parameters. \citet{ref:23} conducted water quality monitoring research in the Zhanghe River basin using UAV remote sensing imagery combined with various machine learning methods (including BP neural networks, Random Forest, XGBoost, etc.), achieving the inversion of key indicators such as chlorophyll $a$, total nitrogen, total phosphorus, and the permanganate index. The results indicated that the stacking ensemble-based machine learning model significantly outperformed single models, improving the accuracy of multi-parameter inversion.

However, these methods are susceptible to various environmental factors in practical applications, with disturbances caused by changes in illumination conditions, specifically the shadow and water surface specular reflection (sun glint) effects, being particularly prominent. The former, primarily caused by shoreline vegetation, buildings, and changes in the UAV's own attitude, can significantly weaken the water body's reflected signal intensity, leading to abnormally low reflectance in local areas. The latter results from strong specular reflection of sunlight on the water surface at specific incidence angles, often masking the true spectral characteristics of the water body \citet{ref:26,ref:27,ref:28,ref:29,ref:30}, causing over-saturation of reflectance in characteristic bands. Moreover, the spatial distribution of shadowed and glint areas is not static; it exhibits significant spatiotemporal dynamics influenced by factors such as changes in the solar elevation angle, cloud drift, surface water disturbances, and wind speed and direction \citet{ref:31,ref:32,ref:33}.

To address this issue, some studies have attempted to mitigate inversion errors caused by illumination differences through post-processing methods such as image radiometric correction, shadow mask segmentation, and reflectance normalization. Other scholars have tried to suppress specular reflection interference using polarization filtering or multi-angle observation techniques \citet{ref:25,ref:28,ref:34,ref:35}. For example, \cite{ref:21} proposed a relative radiometric correction method considering topographic effects for UAV multispectral images, which improved land cover classification accuracy to over 95\% by analyzing the correction's impact on spectral reflectance characteristics. \citet{ref:22} introduced an object-based shadow detection and compensation method for complex urban remote sensing images, achieving accurate shadow detection and effective restoration of shaded area information through an improved Shadow Index (ISI) and Dynamic Penumbra Compensation (DPCM), showing superior performance compared to existing methods. \citet{ref:24} evaluated the impact of atmospheric correction (AC) and sun glint correction on chlorophyll-$a$ (chla) retrieval using Sentinel-2 MSI images in the tropical estuary-lagoon system (MMELS), emphasizing the importance of sun glint and adjacency effect correction. \citet{ref:25} conducted multi-angle water surface sun glint (SG) remote sensing experiments based on multiple UAVs to acquire images under different observation parameters, analyzing the distribution and intensity characteristics of the glint signal, proposing a glint signal extraction model based on regional dark pixel retrieval, and achieving the estimation of water surface roughness and equivalent refractive index by combining the multi-angle glint remote sensing model. This provided a new approach for sun glint remote sensing applications on UAV platforms. However, these methods often rely on static image features, making it difficult to dynamically adapt to complex lighting changes during real-time flight missions. Other studies have attempted to avoid interference areas by pre-setting flight paths, but due to the lack of dynamic sensing and prediction capabilities for shadow and glint areas, their paths are rigid and lagged, making it difficult to maintain monitoring continuity and spectral data quality in time-varying environments \citet{ref:36,ref:37,ref:38}.

Therefore, achieving dynamic prediction and proactive real-time avoidance of light and shadow disturbance areas is a core challenge in improving the effective data acquisition rate and reliability of UAV water quality monitoring. Path planning, as the decision-making center for UAV missions, is key to solving this problem. Traditional path planning algorithms, when dealing with such dynamic, unstructured "soft obstacles," often suffer from insufficient real-time performance or the generation of unsmooth paths, which fails to meet the requirements for continuous and stable flight in water quality monitoring.

To address the challenges mentioned above, this paper proposes a path planning method for UAV water quality monitoring aimed at dynamic light and shadow disturbance avoidance. The core innovations of this research are: first, establishing a dynamic water surface light and shadow prediction model that integrates real-time meteorological and geographical information, abstracting shadow and glint areas into time-varying virtual obstacles; second, designing an improved Interfered Fluid Dynamical System (IFDS) algorithm by introducing a Dynamic Flight Altitude Adjustment (DFAA) mechanism, which incorporates the dynamic obstacle information generated by the prediction model into the three-dimensional path planning in the form of a repulsive force field, enabling the UAV to naturally and smoothly navigate around disturbance areas like fluid flowing around an obstacle; finally, incorporating a Model Predictive Control (MPC) framework combined with a variable Receding Horizon Optimization (RHO) strategy. This allows for proactive decision-making regarding short-term future light and shadow changes while considering flight dynamics constraints, thereby generating a real-time flight path that is energy-optimal and maximizes monitoring efficiency.

\begin{figure}[h]
	\centering
	\includegraphics[width=0.8\textwidth]{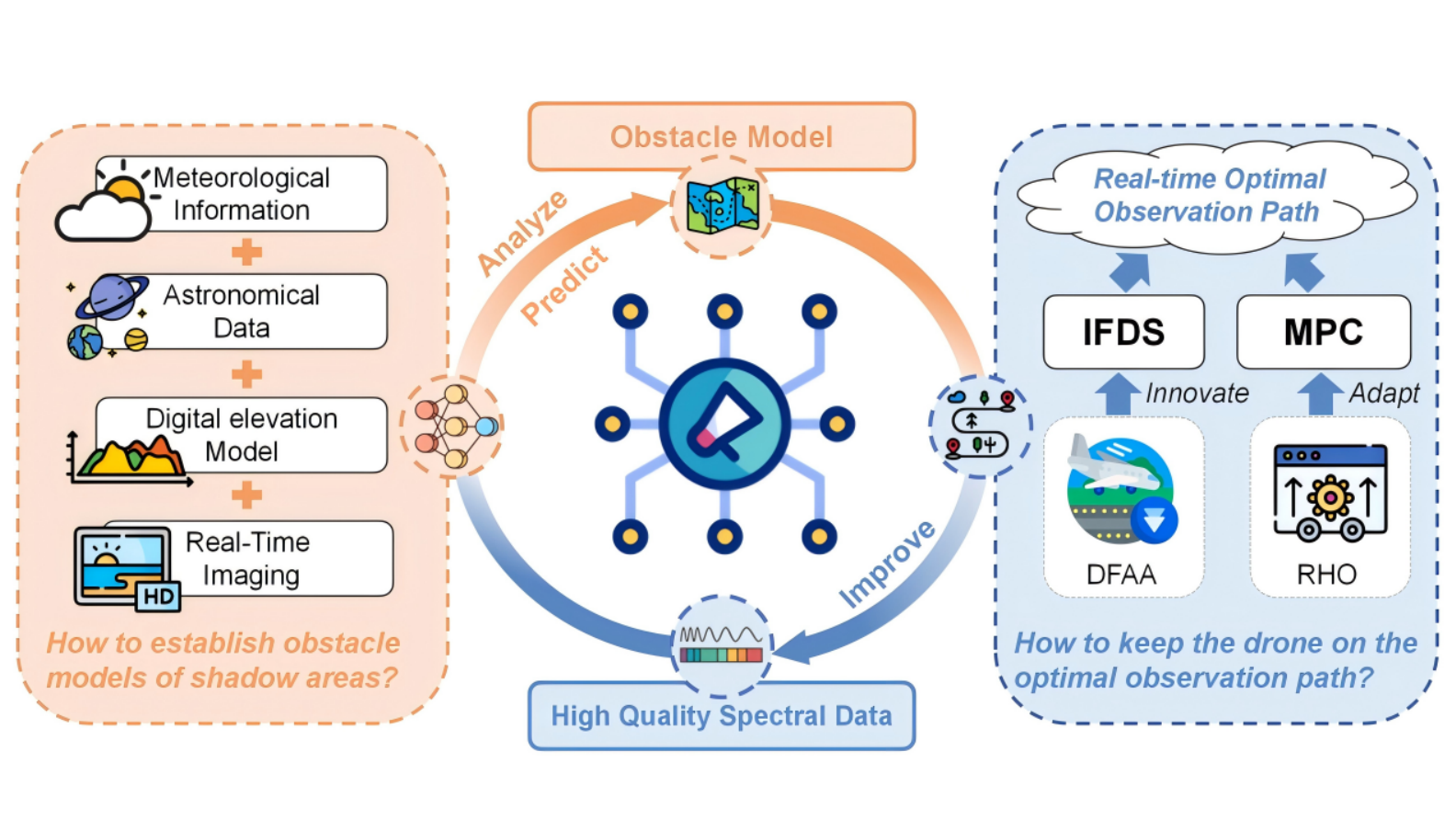}
	\caption{System Architecture Diagram}
	\label{fig:architecture}
\end{figure}

\section{Related Work}

\subsection{Water Quality Detection and Light-Shadow Interference Prediction}
In UAV water quality monitoring tasks, light and shadow interference is one of the primary factors affecting the accuracy of spectral inversion. Due to the high reflectivity and spatiotemporal dynamics of water surfaces, shadows and specular reflections significantly alter the ratio of incident to reflected light. This leads to distortion in the spectral response of key water quality parameters, thereby affecting the stability and generalizability of inversion models. To improve the validity of monitoring data, researchers have begun to establish dynamic prediction models based on the spatial distribution and temporal variation laws of light and shadows, attempting to achieve advance identification and avoidance of interference areas \citet{ref:52,ref:53,ref:54}.

Existing light and shadow prediction methods can be mainly categorized into geometric modeling-based and physical modeling-based approaches. Geometric modeling methods typically utilize solar altitude angle, azimuth angle, and geometric relationships of ground objects to calculate the distribution of shadow areas, which is suitable for static or slowly changing environments. For example, \citet{ref:40} proposed the Aquatic Vegetation Geometric Optical model (AVGO) for discretely distributed aquatic vegetation canopies. By distinguishing between specular and diffuse reflection components of the water background and refining reflection surfaces under different materials and lighting conditions, they achieved accurate simulation of canopy bidirectional reflectance factors in complex scenes using an area-weighted method. Physical modeling methods, on the other hand, combine water surface roughness with incident angles, utilizing specular reflection models or ray-tracing algorithms to simulate the paths of light propagation, refraction, and reflection on the water surface to identify reflective areas. For instance, \cite{ref:41} constructed a bidirectional reflectance model of the water surface based on the physical mechanism of light propagation. By coupling the incident angle, solar altitude angle, and water surface fluctuation characteristics, they accurately characterized the intensity and distribution of specular reflection under different observation geometries, providing a theoretical basis for water body spectral correction under complex lighting conditions. In recent years, some scholars have also attempted to combine time-series image analysis or machine learning methods to extract dynamic features of light and shadow distribution from multi-temporal images, achieving time-series prediction of shadow and reflective areas to provide references for data screening and post-processing radiometric correction. For example, \cite{ref:42} proposed the TSRNet, a shadow removal network based on the combination of language priors and image priors. By introducing natural language descriptions to guide the model in understanding the semantic features of cracks, and combining OTSU segmentation with Bayesian optimization to improve model accuracy, they achieved high-fidelity restoration of crack images in complex shadow environments.

However, existing research mostly focuses on image correction and data quality improvement, lacking studies that combine light and shadow prediction results with UAV path planning or real-time flight control. Especially in complex dynamic environments, the spatial distribution of shadow and reflective areas changes rapidly over time. If prediction-driven path adjustment cannot be achieved, it remains difficult to avoid the impact of light and shadow interference on the continuity and precision of monitoring data. Therefore, there is an urgent need for a method that can fuse dynamic light-shadow prediction with UAV flight strategy optimization to achieve active avoidance of potential interference areas and adaptive task adjustment.

\subsection{UAV Path Planning and Real-time Optimization}
In terms of UAV path planning, to actively avoid shadow and reflective areas that change over time and space, the planned path needs to be dynamically adjusted. Consequently, traditional graph search methods (such as A* and Dijkstra algorithms), although capable of generating globally optimal paths, suffer from high computational complexity and insufficient adaptability to dynamic environments \citet{ref:43,ref:44,ref:55,ref:56}. Rapidly-exploring Random Trees (RRT) and their improved methods accelerate 3D path searching through random sampling, but they often result in poor path smoothness and lack global optimality guarantees \citet{ref:45,ref:57,ref:58,ref:59,ref:60,ref:61}. For instance, \citet{ref:46} proposed an improved Goal-bias RRT algorithm. Through optimized sampling, the introduction of local planning, and node reselection strategies, this method not only shortened the global planning time but also significantly enhanced dynamic obstacle avoidance and path smoothness. However, such methods still mainly rely on local heuristic improvements and lack effective responses to the prediction and active avoidance of time-varying "soft obstacles" like light and shadow interference areas. Intelligent optimization algorithms such as Genetic Algorithms (GA) and Particle Swarm Optimization (PSO) can achieve multi-objective optimization, including path length, energy consumption, and flight altitude, but are still limited in terms of real-time performance and adaptability to dynamic environmental changes \citet{ref:47,ref:48,ref:62,ref:63,ref:64,ref:65}. Addressing these limitations, some studies have attempted to combine path planning with environmental prediction. For example, \citet{ref:49} proposed the Bidirectional Alternating Search A* algorithm (BASA*), which effectively improved the convergence speed of global path planning and path quality through bidirectional alternating expansion and path smoothing. \citet{ref:50} proposed a meta-heuristic optimization algorithm inspired by the behavior of the red-billed blue magpie (RBMO) for numerical optimization, UAV path planning, and engineering design problems, aiming to improve convergence speed, solution accuracy, and robustness, showing excellent performance in 2D/3D path planning. \citet{ref:51} proposed a hybrid Particle Swarm Optimization (PPSwarm) algorithm, which first uses RRT* to generate an initial path, then combines priority planning to assign UAV collaboration order and path randomization strategies. This approach comprehensively considers multi-UAV cooperative constraints and complex environmental obstacles, achieving efficient path search, improved convergence speed, and optimized path quality in obstacle-dense scenarios. This paper proposes an improved Interfered Fluid Dynamical System (IFDS) algorithm combined with flight altitude adjustment, incorporating dynamic obstacle information generated by light-shadow prediction into 3D path planning in the form of a repulsive force field. Furthermore, by combining the Model Predictive Control (MPC) framework with a variable receding horizon optimization strategy, it makes prospective decisions on short-term future light-shadow changes while satisfying flight dynamics constraints, thereby generating a real-time flight path that is energy-optimal and has the highest monitoring efficiency.

In summary, existing research has not yet formed a systematic method that deeply integrates hyperspectral water quality detection, time-series light-shadow prediction, and real-time path optimization. To this end, this study proposes a UAV water quality monitoring path planning scheme oriented towards dynamic light-shadow interference avoidance: in terms of water quality detection, key parameter inversion is achieved based on hyperspectral sensors, while simultaneously predicting water surface shadow and reflective areas in real-time through time-series modeling; in terms of path planning, this paper proposes an adaptive path planning algorithm fusing Interfered Fluid Dynamical System (IFDS) and Model Predictive Control (MPC) to achieve active avoidance of light-shadow interference and global path optimization. This method effectively solves the problem of insufficient adaptability of traditional path planning to dynamic, unstructured light-shadow interference, improving the continuity and data quality of UAV spectral water quality monitoring.

\section{Methodology}
In this section, the shaded areas within the inspection water body are first treated as obstacle threats requiring avoidance during path planning. A mathematical model is established for these obstacles in three-dimensional space, defining parameters such as their position, shape, and dimensions. Next, a random path generation method constructs the drone's initial trajectory and generates an initial velocity field as the target for optimization and modulation. A highly dynamic adjustment mechanism is introduced into the disturbance fluid velocity field to modulate the initial velocity field, achieving obstacle avoidance while maintaining data collection quality. Finally, a rolling optimization strategy is employed to dynamically refine the local path in real-time, ensuring the overall trajectory traversed by the drone achieves optimal performance.

\subsection{Obstacle Constraint Modeling}
This study focuses on the path planning problem of UAVs in dynamic water surface shadow environments, aiming to achieve the dual guarantee of flight safety and observation quality. Water surface shadow areas are formed by the combined effects of terrain undulation, solar altitude angle changes, and cloud distribution. Although they physically appear as two-dimensional distributions, their impact on water quality monitoring tasks performed by UAVs equipped with multispectral cameras specifically manifests as field-of-view occlusion and degradation of perception capabilities \citet{ref:14}. Such impacts can be equivalent to constraints on the feasible region in three-dimensional space during path planning. Therefore, this paper models the dynamic shadow areas as three-dimensional convex obstacles with finite thickness \citet{ref:66} and introduces a path modulation mechanism to achieve active obstacle avoidance and path optimization during the UAV flight process \citet{ref:67}.

\subsubsection{Obstacle Position Modeling}
To accurately describe the spatiotemporal dynamic characteristics of shadow obstacles, utilizing technical means such as digital twins and based on the Digital Elevation Model (DEM), astronomical data (such as solar azimuth), and meteorological information (such as cloud distribution), the state sequence of shadow areas for future moments can be predicted. Let the state of a shadow area at a certain moment be:
\begin{equation}
X_s = (x_s, y_s, z_s, r_s) \label{eq:3-1}
\end{equation}
where $(x_s, y_s, z_s)$ represents the spatial center position of the shadow area, and $r_s$ is its effective action radius in the horizontal direction. To process system noise and improve the robustness of state estimation, the Extended Kalman Filter (EKF) is employed to estimate the shadow state in real-time. Its system model is as follows:
\begin{equation}
\begin{aligned}
X_{k+1} &= f(X_k, u_k) + \omega_k \\
Z_k &= h(X_k) + \upsilon_k
\end{aligned} \label{eq:3-2}
\end{equation}
where $f(\cdot)$ and $h(\cdot)$ are the state transition function and observation function, respectively; $\omega_k \sim N(0, Q)$ and $\upsilon_k \sim N(0, R)$ are the process noise and observation noise, respectively.

\begin{figure}[htbp]
    \centering
    \includegraphics[width=0.6\linewidth]{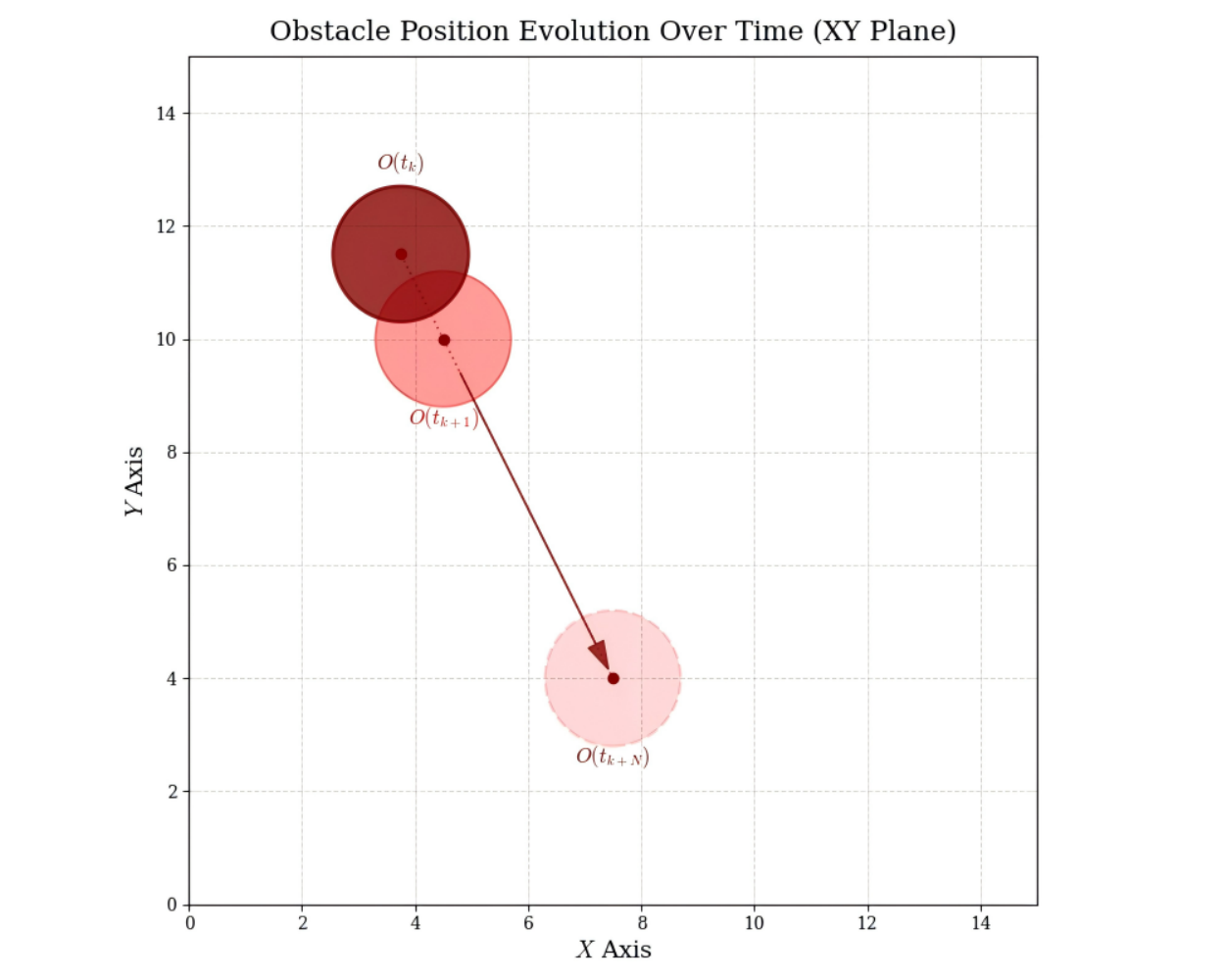}
    \caption{Schematic diagram of obstacle position evolution over time}
    \label{fig:2}
\end{figure}

\subsubsection{Obstacle Geometric Modeling}
To facilitate the integration and implementation of the path planning algorithm, the shadow area at each moment $t$ is modeled as a super-ellipsoid. Its spatial boundary is defined by the following expression:
\begin{equation}
\Gamma(P) = \left(\frac{x-x_0}{\lambda_a a}\right)^{2p} + \left(\frac{y-y_0}{\lambda_b b}\right)^{2q} + \left(\frac{z-z_0}{\lambda_c c}\right)^{2r} \label{eq:3-3}
\end{equation}
where $(x_0, y_0, z_0)$ are the shadow center coordinates; $a$ and $b$ are the effective semi-axis lengths in the horizontal direction; $c$ represents the manually set thickness in the vertical direction, which is usually much smaller than the horizontal scale; $p, q, r \in \mathbb{Z}^+$ are shape control parameters used to adjust the smoothness of the boundary surface. This model constructs an obstacle avoidance volume field with directional perception characteristics and adjustable control force by performing moderate expansion transformation on the two-dimensional shadow area in three-dimensional space.

To further improve the safety margin during the obstacle avoidance process, expansion coefficients $\lambda_a, \lambda_b, \lambda_c > 1$ are further introduced, and Equation (\ref{eq:3-3}) is modified as:
\begin{equation}
\Gamma(P) = \left(\frac{x-x_0}{\lambda_a a}\right)^{2p} + \left(\frac{y-y_0}{\lambda_b b}\right)^{2q} + \left(\frac{z-z_0}{\lambda_c c}\right)^{2r} \label{eq:3-4}
\end{equation}
If a point $P=(x, y, z)$ in space satisfies $\Gamma(P) \le 1$, the point is located inside the shadow obstacle area and is regarded as a non-traversable area for the UAV; when $\Gamma(P) > 1$, the point is located in the feasible space. The above modeling method balances the physical attributes of shadows with the spatial consistency requirements of the path planning algorithm, facilitating the subsequent integration of fluid dynamic modulation, dynamic decision-making, and receding horizon optimization strategies.

\begin{figure}[htbp]
    \centering
    \includegraphics[width=0.8\linewidth]{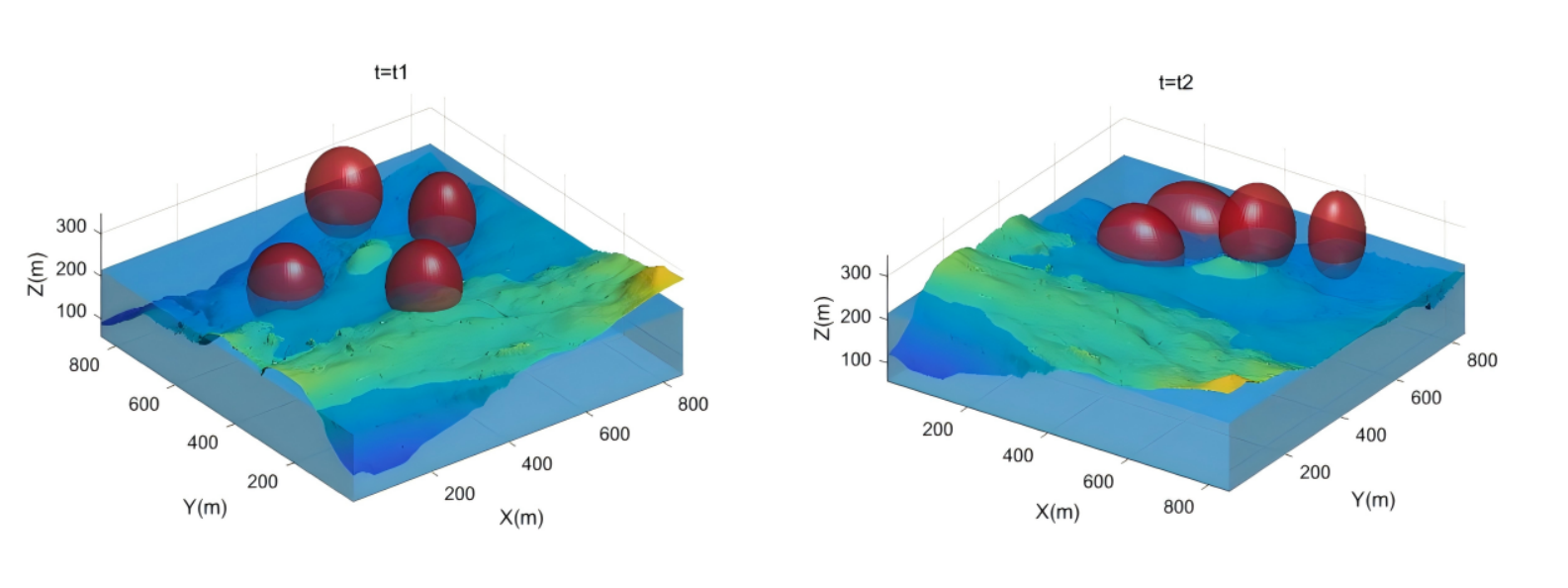}
    \caption{Super-ellipsoid model of dynamic shadow areas}
    \label{fig:3}
\end{figure}

\subsection{Initial Path Construction and Velocity Field Generation}
This paper investigates a real-time path optimization problem in three-dimensional space containing time-dynamic obstacles. To ensure the rationality and executability of the path, it is necessary to consider the maneuverability constraints of the aircraft itself while accounting for obstacle constraints in the environment.

\subsubsection{Random Path Generation}
This paper adopts a random path generation method to construct the starting path as the beginning of the path optimization process. The initial path consists of $K+1$ discrete points, represented as:
\begin{equation}
P = \{P_0, P_1, \dots, P_K\}, \quad P_i = (x_i, y_i, z_i) \label{eq:3-5}
\end{equation}
where $P_0$ and $P_K$ represent the start point and end point of the path, respectively. To guarantee the physical realizability of the path, adjacent path points must satisfy the following maneuverability constraints:
\begin{equation}
\begin{aligned}
|\psi_{i+1} - \psi_i| &\le \omega_{max} \cdot \Delta T \\
\theta_{min} &\le \theta_i \le \theta_{max} \\
h_{min} &\le z_i \le h_{max}
\end{aligned} \label{eq:3-6}
\end{equation}
where $\psi_i$ and $\theta_i$ represent the heading angle and flight path angle at the $i$-th path point, respectively; $\omega_{max}$ is the maximum turn rate; $\Delta T$ is the time interval between path points. $\theta_{max}$ and $\theta_{min}$ are the upper and lower bounds of the flight path angle, respectively; $h_{min}$ and $h_{max}$ are the allowable range of flight altitude.

\subsubsection{Initial Velocity Field Construction}
Based on the obtained sequence of initial path points, the corresponding initial velocity field is constructed through differential operations to serve as the input for the subsequent fluid dynamic modulation. The velocity vector at the $i$-th path point is defined as:
\begin{equation}
u_i = \frac{P_{i+1} - P_i}{\Delta T}, \quad i = 0, 1, \dots, K-1 \label{eq:3-7}
\end{equation}
This velocity field not only describes the expected motion direction of the UAV but also provides a modulated initial flow field for the subsequently introduced Interfered Fluid Dynamical System (IFDS). This maintains the system's goal orientation and dynamic adaptability while ensuring obstacle avoidance capabilities.

\subsection{Construction of Interfered Fluid Velocity Field}
In the actual execution of water quality monitoring tasks by UAVs equipped with multispectral cameras, shadow areas change constantly. To achieve dynamic avoidance of shadow areas by the UAV, this paper introduces the Interfered Fluid Dynamical System (IFDS) to modulate the initial velocity field \cite{ref:68,ref:69}. This method draws on the basic principles of flow around bodies in fluid mechanics. Just as water flows around the outer surface of an obstacle under the action of external forces, the IFDS model treats UAV path points as fluid particles. By constructing an appropriate velocity modulation matrix near the boundary of the obstacle, it guides the path points to deflect tangentially along the obstacle surface, thereby generating a safe and smooth obstacle avoidance path.

Let $u(P)$ be the initial velocity vector at the initial path point $P=(x,y,z)$, and $v_P$ be the relative motion velocity of the shadow area. The modulated velocity field is defined as:
\begin{equation}
u(P) = \beta \cdot M(P) \cdot (u(P) - v_P) + v(P) \label{eq:3-8}
\end{equation}
where $\beta$ is the velocity normalization factor used to maintain the final velocity magnitude at a preset value $V_0$; $M(P) \in \mathbb{R}^{3 \times 3}$ is the modulation matrix used to control the deflection direction and amplitude of the velocity field.

The structure of the modulation matrix is designed as follows:
\begin{equation}
M(P) = I - \omega \cdot n_P n_P^T + \frac{1}{\rho} \cdot \Gamma(P) t(P) n_P^T \label{eq:3-9}
\end{equation}
where $n_P = \nabla \Gamma(P)$ is the outward normal vector of the obstacle boundary at point $P$; $t(P)$ is the tangential guidance direction vector; $\omega_P$ and $\rho$ are the modulation weight and repulsion coefficient, respectively. This construction achieves precise control over the motion direction of path points through the coordinated adjustment of normal and tangential velocity components.

During the path planning process, the system defaults to prioritizing the avoidance of shadow areas in the horizontal direction to maintain optimal flight stability and energy efficiency as much as possible. However, when prediction results indicate that the observable area is too narrow in the horizontal dimension (i.e., the shadow occlusion range is too wide), it may lead to insufficient effective data acquisition. In this case, the system should actively guide the UAV to moderately reduce its flight altitude within the safe altitude range to improve the resolution of sampled images of the effective water surface area, thereby guaranteeing the quality of the water quality monitoring task performed by the UAV.

To this end, this paper introduces a Dynamic Flight Altitude Adjustment (DFAA) mechanism: when the local "effective observable width" $W_{eff}$ is less than the threshold $\tau$ and meets the minimum flight safety conditions, a height guidance perturbation term is added to the modulation matrix:
\begin{equation}
\text{if } W_{eff} < \tau, \quad M_P \leftarrow M_P + \eta \cdot \text{diag}(0, 0, -1) \label{eq:3-10}
\end{equation}
where $\eta > 0$ is the "descent guidance gain," used to control the intensity of altitude adjustment and guide the UAV to lower its flight altitude without affecting safety; the flight altitude constraint $h_{min} \le z_i \le h_{max}$ ensures that the adjustment process always remains within the safe range. This mechanism reflects the mission-oriented path planning concept, which is to prioritize observation quality over mere obstacle avoidance when the feasible space is limited.

\begin{figure}[htbp]
    \centering
    \includegraphics[width=0.8\linewidth]{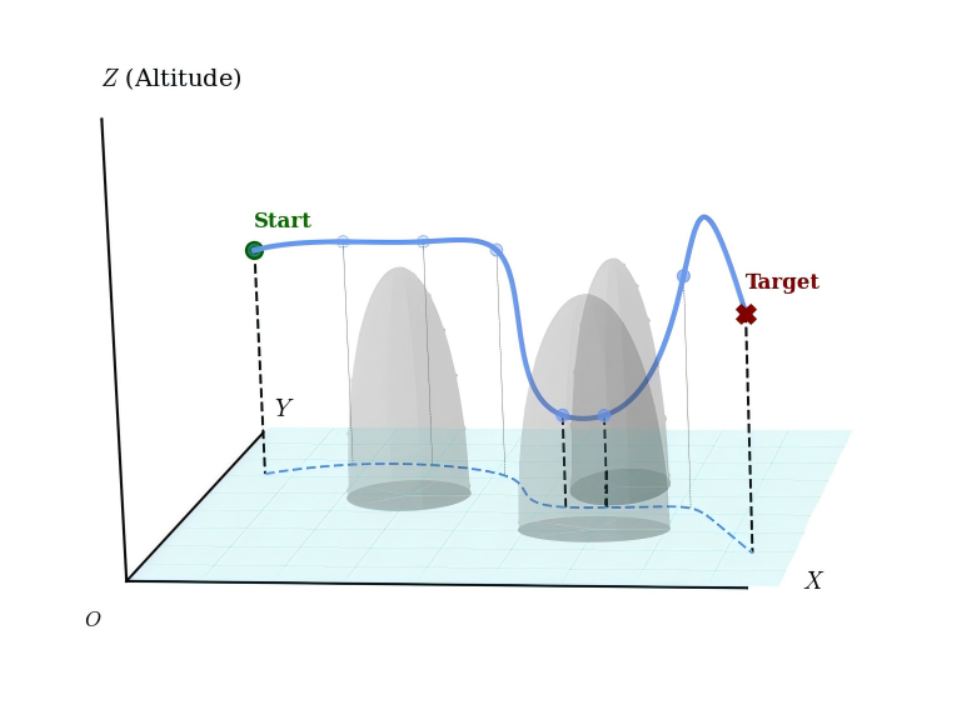}
    \caption{Schematic diagram of the Dynamic Flight Altitude Adjustment mechanism}
    \label{fig:4}
\end{figure}

\subsection{Receding Horizon Optimization Strategy}
To address path feasibility fluctuations caused by the dynamic changes of water surface shadow areas over time, this study adopts a Rolling Optimization (also known as Receding Horizon Optimization, RHO) strategy based on the Model Predictive Control (MPC) framework. Through periodic local path replanning and real-time parameter adjustment, this strategy effectively enhances the UAV's trajectory tracking performance and obstacle avoidance reliability in dynamic environments while ensuring computational real-time performance \citet{ref:70,ref:71,ref:72,ref:73,ref:74}.

\subsubsection{Optimization Window Mechanism}
In each planning cycle, let the UAV's current position be path point $P_k$. The system predicts a local path sequence for the next $N$ steps within its neighborhood:
\begin{equation}
\mathcal{P}_k = \{P_k, P_{k+1}, \dots, P_{k+N}\} \label{eq:3-11}
\end{equation}
By evaluating the feasibility and calculating the cost function value of this sequence, the system selects the optimal control action and executes only the first path point $P_{k+1}$ of the sequence. Subsequently, the optimization window rolls forward by one step length, and prediction and optimization are performed again at the new current position, thereby forming a "look-ahead, execute-one-step" rolling path update mechanism. To guarantee system real-time performance, the selection of the optimization window length $N$ must satisfy $T_p < \Delta T$, where $T_p$ is the time consumption for a single optimization, and $\Delta T$ is the system flight control sampling period.

\begin{figure}[htbp]
    \centering
    \includegraphics[width=0.8\linewidth]{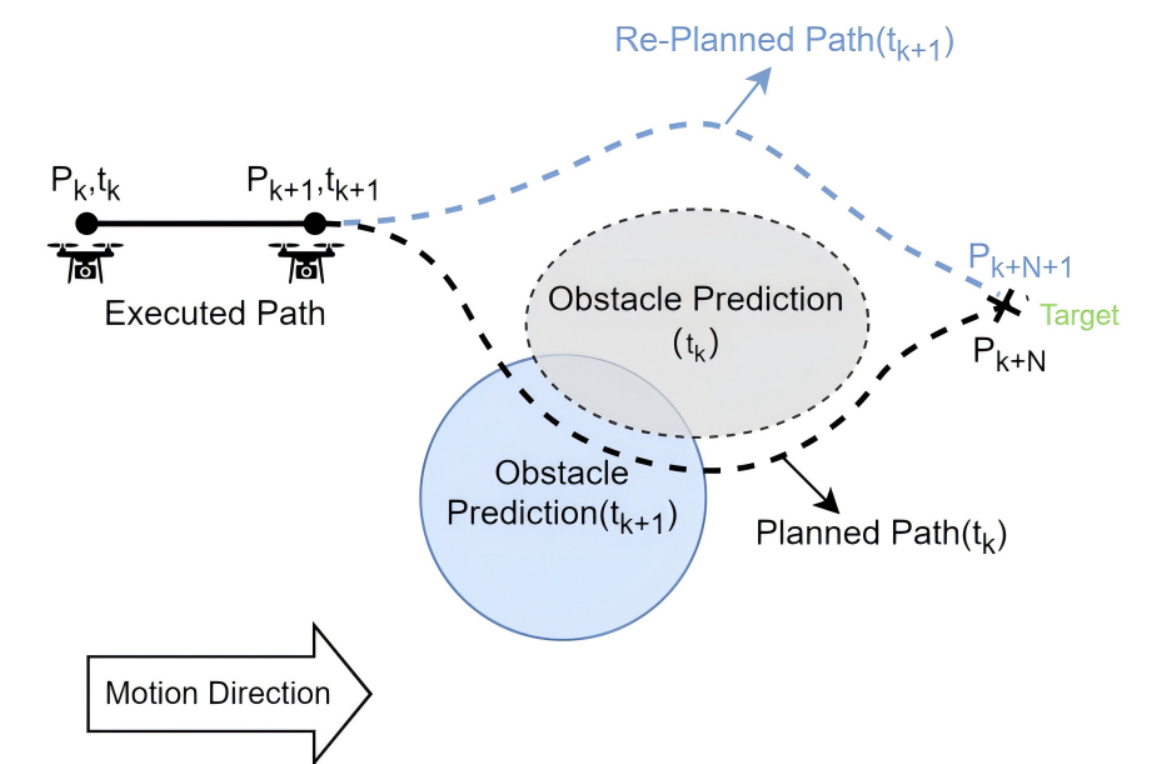}
    \caption{Schematic diagram of the optimization window mechanism}
    \label{fig:5}
\end{figure}

\subsubsection{Path Cost Function Construction}
The goal of rolling optimization is to minimize the following composite cost function while satisfying various constraints:
\begin{equation}
J_k = \lambda_1 T_k + \lambda_2 A_k + \lambda_3 S_k \label{eq:3-12}
\end{equation}
where $T_k$ is the target tracking performance indicator, $A_k$ is the obstacle avoidance penalty term, and $S_k$ is the path smoothness indicator used to measure the drastic degree of changes in heading and inclination; $\lambda_1, \lambda_2, \lambda_3 \in [0, 1]$ are weight coefficients satisfying $\lambda_1 + \lambda_2 + \lambda_3 = 1$.

\paragraph{(1) Target Tracking Term $T_k$}
This term measures the deviation between the interfered path and the original tracking path. Let $u_i$ be the expected velocity vector generated by LGVF at the $i$-th path point, and $u'_i$ be the actual velocity vector modulated by IFDS, then:
\begin{equation}
T_k = \sum_{i=k}^{k+N-1} (1 - \cos\langle u_i, u'_i \rangle), \quad \langle u_i, u'_i \rangle = \angle(u_i, u'_i) \label{eq:3-13}
\end{equation}
A smaller value indicates that the interfered path is closer to the original tracking vector, and the target tracking performance is better.

\paragraph{(2) Obstacle Penalty Term $A_k$}
This term measures the minimum distance between path points and obstacle boundaries to prevent the path from intruding into shadow areas. Let the boundary function of the $w$-th shadow obstacle body be $\Gamma_w(P)$, then define:
\begin{equation}
A_k = \sum_{w=1}^{W} \sum_{i=k}^{k+N} \phi(\Gamma_w(P_i)) \label{eq:3-14}
\end{equation}
where the penalty function $\phi(\cdot)$ is defined in a piecewise form as follows:
\begin{equation}
\phi(\Gamma) = \begin{cases} 
0, & \Gamma > \gamma_{safe} \\ 
\infty, & \Gamma \le 1 \\ 
\frac{1}{\Gamma-1}, & 1 < \Gamma \le \gamma_{safe} 
\end{cases} \label{eq:3-15}
\end{equation}
where $\gamma_{safe} > 1$ is a preset safety distance threshold.

\paragraph{(3) Path Smoothness Term $S_k$}
The smoothness term is used to penalize drastic angle changes in the trajectory to maintain dynamic executability. Let the heading angle and flight path angle between path points be $\psi_i, \theta_i$ respectively, then:
\begin{equation}
S_k = \sum_{i=k}^{k+N-1} (\mu_1 |\psi_{i+1} - \psi_i| + \mu_2 |\theta_{i+1} - \theta_i|) \label{eq:3-16}
\end{equation}
where $\mu_1, \mu_2$ are smoothness weight coefficients, reflecting the emphasis on lateral and longitudinal angular continuity.

\subsubsection{Parameter Optimization and Feasibility Judgment}
All candidate paths must satisfy the aircraft's dynamic and environmental constraints, including but not limited to:
\begin{itemize}
    \item Maximum turn rate constraint $|\psi_{i+1} - \psi_i| \le \omega_{max}\Delta T$;
    \item Flight path angle constraint $\theta_{min} \le \theta_i \le \theta_{max}$;
    \item Flight altitude constraint $h_{min} \le z_i \le h_{max}$;
    \item Shadow avoidance constraint $\Gamma_w(P_i) > 1, \forall i, w$.
\end{itemize}
Paths that fail to satisfy any of the above constraints will be immediately discarded, and their corresponding cost function $J(k)$ is regarded as infinity.

To obtain the optimal path, the system optimizes the IFDS parameter set within each planning cycle through enumeration or heuristic search:
\begin{equation}
U^* = \mathop{\arg\min}_{U} J(k)
\end{equation}
In each planning cycle, the optimization process executes only the first step point $P_{k+1}$ of the selected optimal path and updates the path start index $k \leftarrow k+1$. When the UAV reaches the target position or the path has been planned to the end point, the algorithm terminates. This rolling optimization mechanism, through the combination of local iterative optimization and global feasibility guarantees, enables the UAV to achieve safe, smooth, and task-consistent trajectory tracking in dynamic shadow environments.

\subsection{Convergence Analysis}
To verify the stability and trajectory convergence of the proposed path planning method in dynamic shadow environments, this section conducts a theoretical analysis based on Lyapunov stability theory \citet{ref:75}. The core idea is to define a Lyapunov candidate function that reflects the deviation between the UAV and the target observation state. By combining the modulation characteristics of the Interfered Fluid Dynamical System (IFDS) with the rolling optimization mechanism of Model Predictive Control (MPC), it is proven that the closed-loop system trajectory eventually converges to the preset optimal observation region and maintains stability under dynamic shadow constraints.

\subsubsection{System State and Target Observation Model Definition}
First, the core control objective of the UAV is clarified: to maintain the optimal lateral orbiting radius and the best observation altitude while avoiding dynamic shadows, so as to guarantee the spectral data quality of water quality monitoring. Define the following key state variables:

Let the current position of the UAV be $P=(x,y,z)$ and the preset tracking target position be $P_t=(x_t,y_t,z_t)$. The lateral distance $r$ between the UAV and the target on the horizontal plane is defined as:
\begin{equation}
r = \sqrt{(x-x_t)^2 + (y-y_t)^2} \label{eq:3-18}
\end{equation}

The height deviation $h$ between the UAV and the target in the vertical direction is defined as:
\begin{equation}
h = z - z_t \label{eq:3-19}
\end{equation}

The ideal convergence state of the system is for the UAV to maintain $r=R$ and $h=H$, i.e., converging to the target set:
\begin{equation}
S_0 = \{(r,h) \mid r=R, h=H\} \label{eq:3-20}
\end{equation}
where $R$ is the expected lateral orbiting radius; $H$ is the optimal observation altitude, which must satisfy two constraints: first, it must be higher than the safe altitude of the water surface to avoid airflow disturbances; second, it must be lower than the maximum allowable altitude to ensure the intensity of spectral reflectance signals, i.e., $h_{min} \le z_t+H \le h_{max}$.

\subsubsection{Lyapunov Candidate Function Construction}
The core of Lyapunov stability theory is to determine a positive definite function whose derivative is non-positive, thereby proving that the system deviation will gradually decay to zero. Combined with the state definitions in this section, the candidate Lyapunov function is constructed as follows:
\begin{equation}
V(P) = \frac{1}{2}(r^2 - R^2)^2 + \frac{1}{2}(h^2 - H^2)^2 \label{eq:3-21}
\end{equation}

When the UAV is in the ideal observation state, i.e., $r=R$ and $h=H$, $V(P) = 0$; when the UAV deviates from the ideal state, i.e., $r \neq R$ or $h \neq H$, $(r^2 - R^2)^2 > 0$ or $(h^2 - H^2)^2 > 0$. Therefore, this Lyapunov candidate function satisfies the positive definite condition and can be used to characterize the "energy level" of the system deviation.

\subsubsection{Key Assumptions and System Characteristic Analysis}
To ensure the rigor of the convergence analysis, it is necessary to clarify the core assumptions satisfied by the system. The following assumptions are based on the design logic of the proposed method and possess physical realizability:

\paragraph{Assumption 1: Feasible Safe Domain Constraint}
The motion state of the UAV is always located within a compact obstacle-free safe set $X_{safe} \cap C_{free}$. Here, $X_{safe}$ is the flight envelope constraint set, satisfying the maximum turn rate $|\psi| \le \omega_{max}$, maximum climb angle $|\theta| \le \theta_{max}$, and altitude constraint $h_{min} \le z \le h_{max}$; $C_{free}$ is the feasible region without shadow interference, meaning the path points satisfy the boundary condition $\Gamma(P) > 1$ of the shadow obstacle bodies. This assumption ensures that the UAV will not cause system instability due to exceeding dynamic limits or shadow interference.

\paragraph{Assumption 2: "Non-counter-flow" Characteristic of IFDS Modulation Matrix}
The IFDS modulation matrix $M(P) \in \mathbb{R}^{3 \times 3}$ needs to satisfy the "non-counter-flow" condition:
\begin{equation}
u(P) \cdot u^*(P) \ge 0 \label{eq:3-22}
\end{equation}
where $u(P) = \beta \cdot M(P) \cdot (u_0(P) - v(P)) + v(P)$ is the actual velocity vector after IFDS modulation; $u^*(P) \triangleq -k \nabla V(P), k>0$ is the virtual guidance velocity vector. This vector is only used for analysis and does not participate in actual control; its direction points to the ideal observation state, guiding the UAV to converge towards $r=R, h=H$.

The physical meaning of the "non-counter-flow condition" is that IFDS modulation will not cause the UAV's motion direction to deviate from the ideal observation state, ensuring that the convergence trend towards the target is maintained during the obstacle avoidance process. Meanwhile, the modulation matrix $M(P)$ is continuously bounded, avoiding trajectory oscillations caused by sudden velocity changes.

\paragraph{Assumption 3: Stability of Vertical Height Adjustment}
When the effective observable width $W_{eff}$ in the horizontal direction is less than the threshold $\tau$ and meets the minimum safety conditions, the system activates vertical height adjustment. At this time, the velocity perturbation term in the height direction $\delta u = [0, 0, \delta u_z]^T$ needs to satisfy:
\begin{equation}
\delta u_z \propto -h(h^2 - H^2) \label{eq:3-23}
\end{equation}
The design logic of this perturbation term is: when the UAV is higher than the target altitude, i.e., $h>0$, then $\delta u_z < 0$, guiding the UAV to descend; when the UAV is lower than the target altitude, i.e., $h<0$, then $\delta u_z > 0$, guiding the UAV to ascend. This ensures that the height adjustment process still converges towards the target. By applying velocity saturation processing, the velocity magnitude $|u+\delta u| = V_0$ is maintained without violating maneuverability constraints such as attitude and turn rate.

\paragraph{Assumption 4: Recursive Feasibility of MPC Rolling Optimization}
The rolling optimization window of MPC needs to satisfy recursive feasibility: if a feasible path satisfying all constraints exists at the current moment, a feasible path still exists after the optimization window rolls forward at the next moment. This is specifically achieved through the following designs:
\begin{itemize}
    \item A terminal cost term is introduced into the cost function to impose a greater penalty on the deviation of path points at the end of the window, guiding the path to extend towards a long-term feasible region;
    \item The optimization window length $N$ satisfies $T_p < \Delta T$, ensuring real-time performance while reserving sufficient optimization space. Here, $T_p$ is the time consumption for a single optimization, and $\Delta T$ is the control sampling period.
\end{itemize}

\subsubsection{Convergence Theorem and Proof}
Based on the above definitions and assumptions, the following system convergence theorem is proposed:

\textbf{Theorem: Trajectory Convergence in Dynamic Shadow Environments}

If the system satisfies Assumptions 1-4, and the closed-loop guidance law is:
\begin{equation}
u_{total}(P) = \beta M(P)(u_0(P) - v(P)) + v(P) + \delta u \label{eq:3-24}
\end{equation}
where $u_0(P)$ is the initial velocity field obtained by random path differencing, $v(P)$ is the relative motion velocity of the shadow obstacle, and $|u_{total}(P)| = V_0$. Then the following conditions hold:
\begin{enumerate}
    \item The UAV's closed-loop trajectory $P(t)$ is globally bounded and always located within the obstacle-free safe set $X_{safe} \cap C_{free}$;
    \item The trajectory asymptotically converges to the optimal observation position set $S = S_0 \cap X_{safe} \cap C_{free}$;
    \item If exact convergence to $S$ is impossible due to shadow or altitude constraints, the trajectory converges to the $\varepsilon$-neighborhood of $S$, where $\varepsilon$ is determined by the tightness of the constraints.
\end{enumerate}

\textbf{Proof:}

According to the chain rule, the derivative of the Lyapunov function $V(P)$ with respect to time is:
\begin{equation}
\dot{V} = \frac{\partial V}{\partial P} \cdot \dot{P} = [\nabla V(P)]^T \cdot (u(P) + \delta u) \label{eq:3-25}
\end{equation}
where $\dot{P} = u(P) + \delta u$ is the actual motion velocity of the UAV containing the height adjustment perturbation term. The gradient $\nabla V(P)$ reflects the direction of change in deviation energy.

From $V(P) = \frac{1}{2}(r^2 - R^2)^2 + \frac{1}{2}(h^2 - H^2)^2$, combined with $r = \sqrt{(x-x_t)^2 + (y-y_t)^2}$ and $h = z - z_t$, calculating the partial derivatives yields:
\[ \nabla V(P) = \left[ 2r(r^2 - R^2) \cdot \frac{x-x_t}{r}, 2r(r^2 - R^2) \cdot \frac{y-y_t}{r}, 2h(h^2 - H^2) \right]^T \]

Simplifying, we get:
\begin{equation}
\nabla V(P) = [2(r^2 - R^2)(x-x_t), 2(r^2 - R^2)(y-y_t), 2h(h^2 - H^2)]^T \label{eq:3-26}
\end{equation}

The core of convergence is to prove that the deviation energy does not increase, i.e., $\dot{V} \le 0$, and $\dot{V}=0$ only when $P \in S$. The discussion is divided into two scenarios below.

(i) When the vertical guidance mode is not triggered, at this time $\delta u = 0$, and the closed-loop velocity is $u(P)$. According to the "non-counter-flow property" of Assumption 2, $u(P) \cdot u^*(P) \ge 0$, and since $u^*(P) = -k \nabla V(P), k>0$, we obtain:
\begin{equation}
u(P) \cdot \nabla V(P) \le 0 \label{eq:3-27}
\end{equation}
Substituting into Equation (\ref{eq:3-25}), we get:
\begin{equation}
\dot{V} = (\nabla V)^T \cdot u \le 0 \label{eq:3-28}
\end{equation}
The equality holds if and only if $u(P) \cdot \nabla V(P) = 0$. Combined with the gradient expression in Equation (\ref{eq:3-26}), this condition corresponds to the following two cases:
\begin{itemize}
    \item $\nabla V = 0$, i.e., $r^2 - R^2 = 0$ and $h^2 - H^2 = 0$, at which point the UAV reaches the ideal observation state $P \in S_0$;
    \item $u(P)$ is perpendicular to $\nabla V(P)$. However, due to the non-counter-flow characteristic of IFDS, $u(P)$ always contains a convergence component towards $S_0$, so this case is only a transient state and will not be maintained for a long time.
\end{itemize}
Therefore, when height adjustment is not activated, $\dot{V} \le 0$, and $\dot{V}=0$ only when $P \in S_0$.

(ii) When the shadow triggers the vertical guidance mode, the height adjustment perturbation term $\delta u$ needs to be considered. Equation (\ref{eq:3-25}) becomes:
\begin{equation}
\dot{V} = (\nabla V)^T \cdot u + (\nabla V)^T \cdot \delta u \label{eq:3-29}
\end{equation}
It has been proven in scenario (i) that $(\nabla V)^T \cdot u \le 0$. It only remains to prove $(\nabla V)^T \cdot \delta u \le 0$. According to Assumption 3, $\delta u = [0, 0, \delta u_z]^T$ and $\delta u_z \propto -h(h^2 - H^2)$. Taking the proportionality coefficient as a negative constant $-k_h (k_h > 0)$, then $\delta u_z = -k_h h(h^2 - H^2)$. Substituting the vertical component of the gradient $\nabla V_z = 2h(h^2 - H^2)$, we get:
\begin{equation}
\nabla V_z \cdot \delta u_z = 2h(h^2 - H^2) \cdot [-k_h h(h^2 - H^2)] = -2k_h h^2(h^2 - H^2)^2 \le 0 \label{eq:3-30}
\end{equation}
Therefore, after activating height deviation adjustment, similarly $\dot{V} \le 0$, and $\dot{V}=0$ only when $P \in S_0$.

According to LaSalle's Invariance Principle: if the Lyapunov function $V(P)$ is positive definite, its derivative $\dot{V}(P) \le 0$, and the largest invariant set within $V(P)=0$ contains only the target set $S$, then the system trajectory will asymptotically converge to $S$.

Based on the above analysis, the following conclusions can be drawn:
\begin{itemize}
    \item The set where $\dot{V}(P)=0$ is $\{P \mid \nabla V=0\} \cap X_{safe} \cap C_{free} = S$;
    \item This set is an invariant set. Once the UAV enters $S$, $\dot{V}=0$, and it will always remain within $S$.
\end{itemize}
Therefore, the UAV trajectory will asymptotically converge to $S$. If it cannot precisely reach $S$ due to shadow occlusion or altitude constraints, the trajectory will converge to the $\varepsilon$-neighborhood of $S$.

From Assumption 1, the UAV is always located within the compact set $X_{safe} \cap C_{free}$, and $V(P)$ is positive definite and bounded (since $r$ and $h$ are both bounded). Also, because $\dot{V} \le 0$, $V(P)$ is non-increasing over time, so $V(P(t)) \le V(P(0))$, meaning $r(t)$ and $h(t)$ are always bounded. Thus, the closed-loop trajectory $P(t)$ is globally bounded.

In summary, $V(P)$ constitutes a common Lyapunov function for the system under both modes. $\dot{V} \le 0$ holds constantly, and the largest invariant set for $\dot{V}=0$ is $S$ (or its constrained subset). Therefore, under unconstrained conditions, the closed-loop system is asymptotically stable with respect to $S$; under constrained conditions, the system still possesses practical stability and converges to the target neighborhood.

\section{Simulation}

To systematically verify the effectiveness of the proposed dynamic light-shadow interference avoidance path planning method (IFDS+MPC) for UAV water quality monitoring tasks, three progressive simulation experiments were designed: basic simulation experiments aim to verify the core obstacle avoidance performance, observation quality, and path optimization effects; DFAA mechanism ablation simulation focuses on verifying the improvement of observation data quality by the Dynamic Flight Altitude Adjustment (DFAA) mechanism; robustness tests verify the algorithm's stable adaptability in real complex scenarios by introducing environmental perception noise. All simulations were built based on the Python platform, combining multi-dimensional performance indicators and visualization results to comprehensively evaluate the superiority and practicality of the algorithm.

\subsection{Simulation Environment and Parameter Settings}
The simulation scenario takes a simulated river channel as the monitoring object, using Spline curves to generate the river contour, and characterizing time-varying shadows and dynamic obstacles through super-ellipsoid or polygonal prism models. The UAV dynamics model is simplified based on a six-degree-of-freedom rigid body model, considering physical constraints in actual flight. During the simulation, the initial path, river centerline, and camera viewing angle parameters are configured, outputting snapshots in PNG format (trajectory, coverage range) and data in CSV format for subsequent quantitative analysis.

\begin{figure}[htbp]
    \centering
    \includegraphics[width=1.0\linewidth]{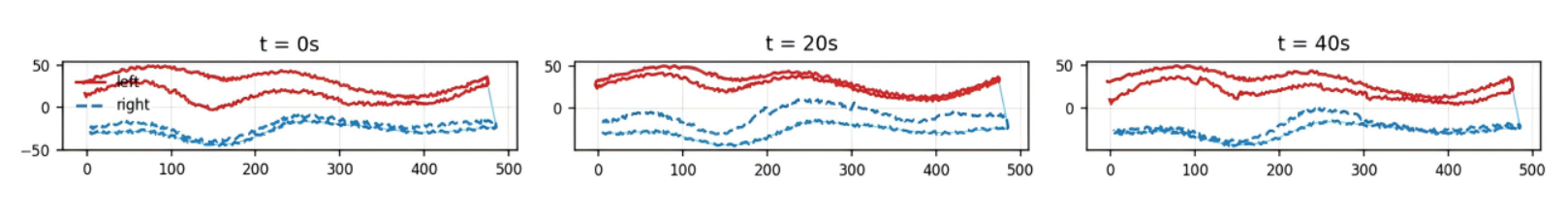}
    \caption{Distribution of water surface shadows at different moments}
    \label{fig:6}
\end{figure}

Key simulation parameter settings are shown in Table \ref{tab:1}. All parameters are based on typical configurations of actual UAV water quality monitoring tasks to ensure the authenticity and reliability of the simulation. The spatial distribution of water surface shadows at different moments is shown in Figure \ref{fig:6}, where the boundaries exhibit significant fluctuations and spatiotemporal dynamics, providing a realistic test scenario for dynamic obstacle avoidance verification.

\begin{table}[htbp]
    \centering
    \caption{Key Parameter Settings for Simulation Experiments}
    \label{tab:1}
    \begin{tabular}{lcc}
        \hline
        \textbf{Parameter Name} & \textbf{Symbol} & \textbf{Value/Unit} \\
        \hline
        System Sampling Period & $\Delta T$ & 0.1 s \\
        UAV Cruise Speed & $V_0$ & 10 m/s \\
        Max Turn Rate & $\omega_{max}$ & 0.5 rad/s \\
        Safety Distance Threshold & $\gamma_{safe}$ & 1.2 \\
        MPC Prediction Horizon & $N$ & 20 \\
        IFDS Repulsion Coefficient & $\rho$ & 1.5 \\
        Observation Width Threshold & $W_{eff}^{th}$ & 30 m \\
        \hline
    \end{tabular}
\end{table}

To effectively handle these dynamic shadow areas during path planning, it is necessary to convert them into geometric representations that facilitate mathematical modeling and optimization solutions. To this end, principal component analysis (PCA) was employed to transform the dynamic shadow data in Figure \ref{fig:6} into an equivalent hyperellipsoidal model. Specifically, by analyzing the shadow distribution data at time $t = 0s$, 7 to 8 ellipsoids of different sizes and orientations are extracted. These ellipsoids can effectively approximate the geometric characteristics of the original shadow areas. Each ellipsoid is parameterized by its center position, principal axis direction, and semi-axis length, providing standardized obstacle constraints for subsequent trajectory optimization algorithms. More importantly, during the Rolling Horizon Optimization process based on Model Predictive Control (MPC), geometric parameters such as the center position, principal axis direction, and semi-axis length of these super-ellipsoids are updated in real-time according to the dynamic shadow observation data at the current moment. This enables the UAV to make timely and effective responses to the time-varying nature of obstacles. This conversion process from original shadow data to geometric models not only maintains the dynamic characteristics of shadow areas but also significantly simplifies the mathematical expression of collision detection and obstacle avoidance constraints in trajectory planning, laying the foundation for MPC-based trajectory optimization.

\begin{figure}[htbp]
    \centering
    \includegraphics[width=0.7\linewidth]{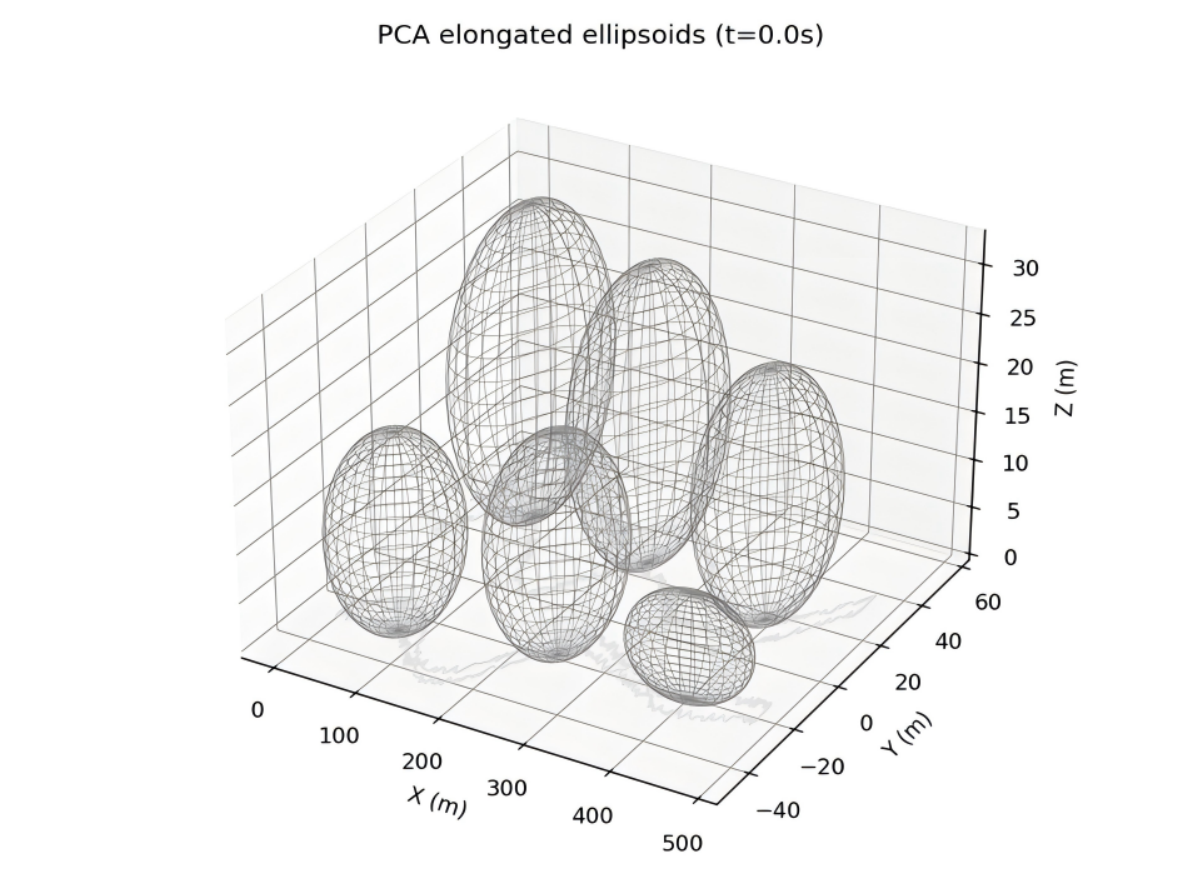}
    \caption{Dynamic obstacle equivalent super-ellipsoid model of water surface shadows}
    \label{fig:7}
\end{figure}

\subsection{Basic Simulation Experiment}

\subsubsection{Simulation Steps}
According to the method designed previously, the path reference field (such as LGVF or centerline tangent) provides the trend of forward propulsion and normal correction. Combined with the implicit function of obstacles and the IFDS modulation matrix constructed based on normal and tangential directions, core regulation is achieved. The "shadow-free ratio" is used as an indicator to achieve smooth self-adaptation of flight altitude through PID. Simultaneously, a baseline scheme of "Pure PID lateral control + Simple obstacle avoidance" is set up, where the "simple obstacle avoidance" mechanism adopts an immediate rule based on boundary detection, triggering simple steering or deceleration actions only when obstacles enter the warning range. Furthermore, steering bias and velocity scale can be optimized in the rolling time domain through MPC to minimize tracking, obstacle avoidance, and smoothing costs. Regarding the mathematical model and control flow, UAV velocity and acceleration limits and discrete integration are considered. The effective coverage area and ratio are calculated based on the Field of View (FoV) observation geometry, and a cost function containing "tracking error," "obstacle penalty," and "smoothing term (heading change rate)" is constructed.

\subsubsection{Analysis of Basic Simulation Results}
\begin{figure}[htbp]
    \centering
    \includegraphics[width=0.9\linewidth]{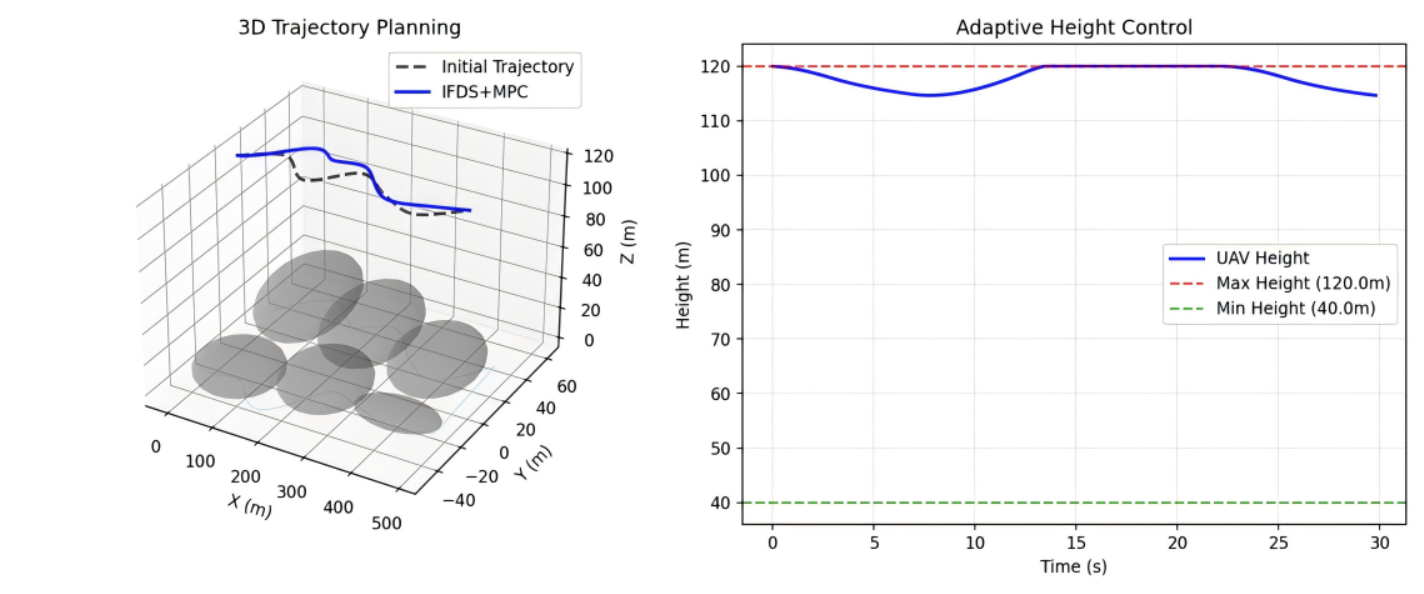}
    \caption{Optimal path formed based on IFDS+MPC in simulation}
    \label{fig:8}
\end{figure}

Based on the obstacle model established in Figure \ref{fig:7}, Figure \ref{fig:8} further verifies the effectiveness of the proposed IFDS+MPC hybrid control strategy. By comparing the initial trajectory (gray dashed line) and the optimized trajectory (blue solid line), it is found that the optimized trajectory can intelligently avoid all super-ellipsoid obstacles while maintaining good path smoothness and safety. Especially in obstacle-dense areas, the trajectory achieves effective obstacle avoidance through precise lateral and altitude adjustments, fully demonstrating the synergistic effect of IFDS local obstacle avoidance capability and MPC global optimization characteristics.

\begin{figure}[htbp]
    \centering
    \includegraphics[width=1.0\linewidth]{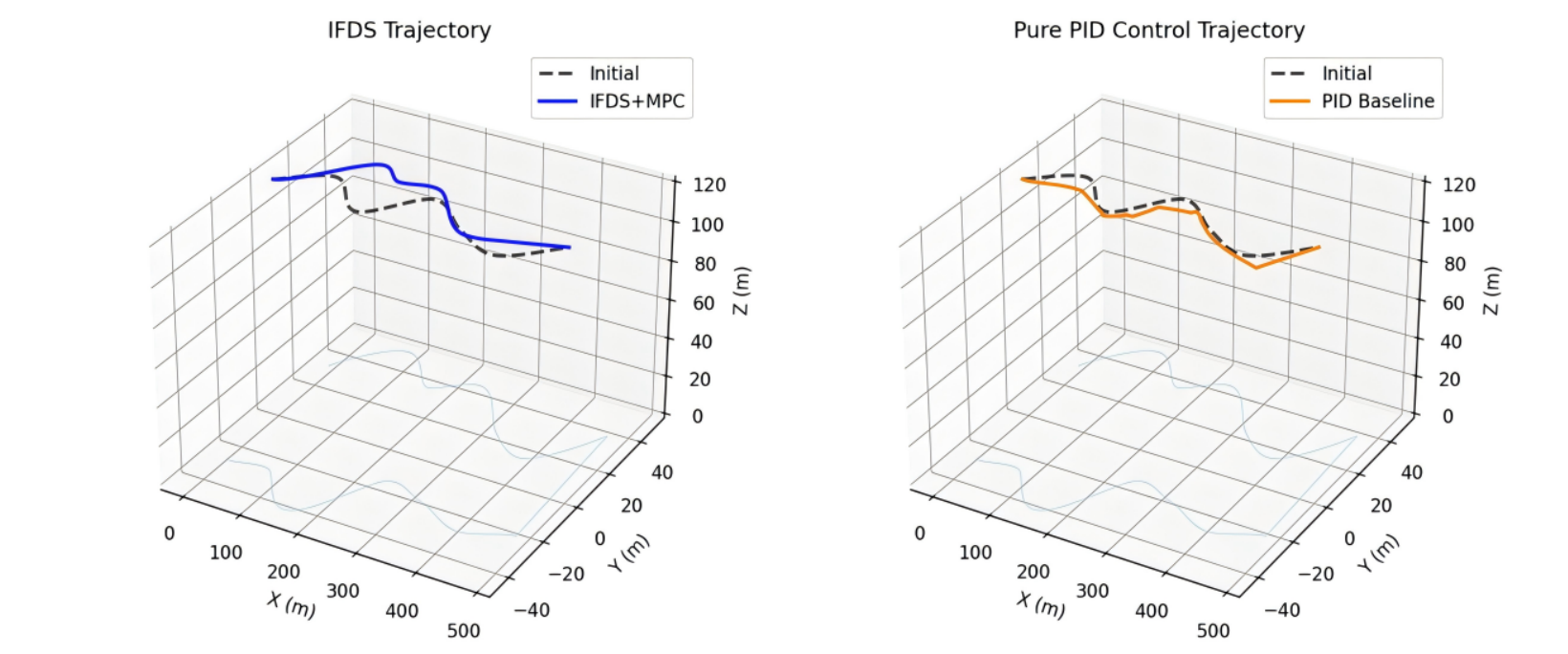}
    \caption{Comparison of trajectory planning results between IFDS+MPC method and PID method}
    \label{fig:9}
\end{figure}

The optimized trajectory can intelligently avoid obstacles while the altitude control strictly maintains the safety range of 40-120m, reflecting good environmental adaptability and control precision. To further verify the superiority of the proposed method, Figure \ref{fig:9} compares the IFDS+MPC strategy with the traditional PID control method. The comparison shows that the IFDS+MPC trajectory has smoother path characteristics and more effective obstacle avoidance capabilities compared to the PID baseline method. Although PID control can also achieve basic trajectory tracking, it shows obvious limitations when dealing with complex obstacle environments, which further highlights the technical advantages of the IFDS+MPC hybrid strategy in dynamic environments.

The performance difference between the IFDS+MPC method and the pure PID control method can be evaluated through the following three dimensions of comparative analysis.
\textbf{Trajectory Quality:} Although PID control can achieve basic trajectory tracking, it exhibits more sharp turns and path discontinuity problems when dealing with complex obstacle environments, primarily due to its lack of global optimization capabilities and adaptability to dynamic environments.
\textbf{Observation Performance:} From the two indicators of effective coverage area and coverage ratio, the IFDS+MPC method is significantly superior to PID control in observation performance. In terms of effective coverage area, the peak value of the IFDS+MPC method reaches about $7000 m^2$, while the PID method is only $5500 m^2$, an increase of about 27\%. In terms of coverage ratio, the IFDS method maintains a stable level of 0.45-0.5, while the PID method fluctuates between 0.4-0.45, with relatively poor stability. This indicates that within the same flight time, the IFDS+MPC method can obtain more comprehensive and continuous water quality data, providing more reliable data support for environmental monitoring.
\textbf{Dynamic Obstacle Avoidance:} Facing time-varying shadow areas in the river channel, the IFDS+MPC method achieves real-time response to dynamic obstacles through rolling horizon optimization, while PID control shows obvious lag and insufficient adaptability. This is of great value in tasks requiring high real-time performance such as water quality monitoring.

\begin{figure}[htbp]
    \centering
    \includegraphics[width=0.8\linewidth]{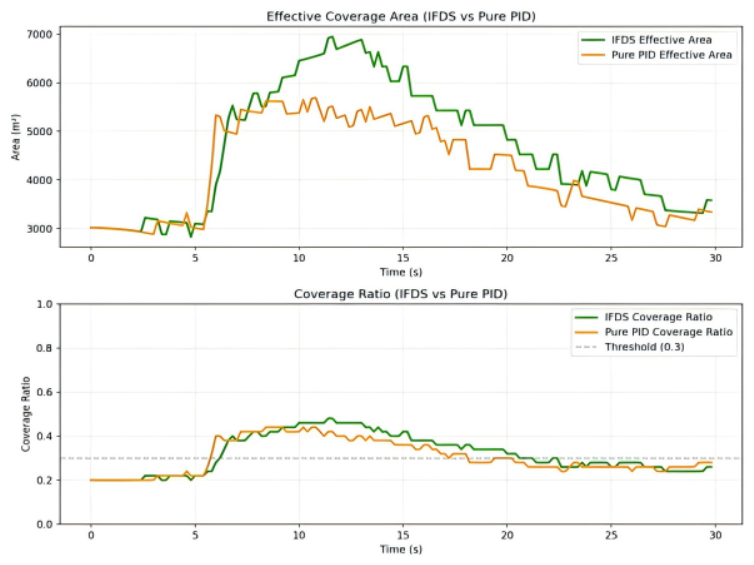}
    \caption{Comparison of effective coverage area and coverage ratio between IFDS+MPC method and pure PID control method}
    \label{fig:10}
\end{figure}

The simulation results show that the IFDS+MPC hybrid strategy, by combining local obstacle avoidance capabilities and global optimization characteristics, not only achieves smoother trajectory planning but also demonstrates significant performance advantages in observation task execution, verifying the effectiveness and practicality of the proposed method in complex dynamic environments.

\subsubsection{Algorithm Performance Evaluation in Multi-Scenarios}
To comprehensively verify the effectiveness of the algorithm under different actual working conditions and overcome the limitations of single-scenario testing, we constructed two typical water quality monitoring scenarios: "Sparse Interference" and "Dense Interference," and conducted 50 Monte Carlo simulation experiments for each. The comparison algorithms include: (1) Traditional PID tracking control; (2) Reactive obstacle avoidance with IFDS only; (3) The IFDS+MPC hybrid strategy proposed in this paper.

\begin{figure}[htbp]
    \centering
    \includegraphics[width=0.8\linewidth]{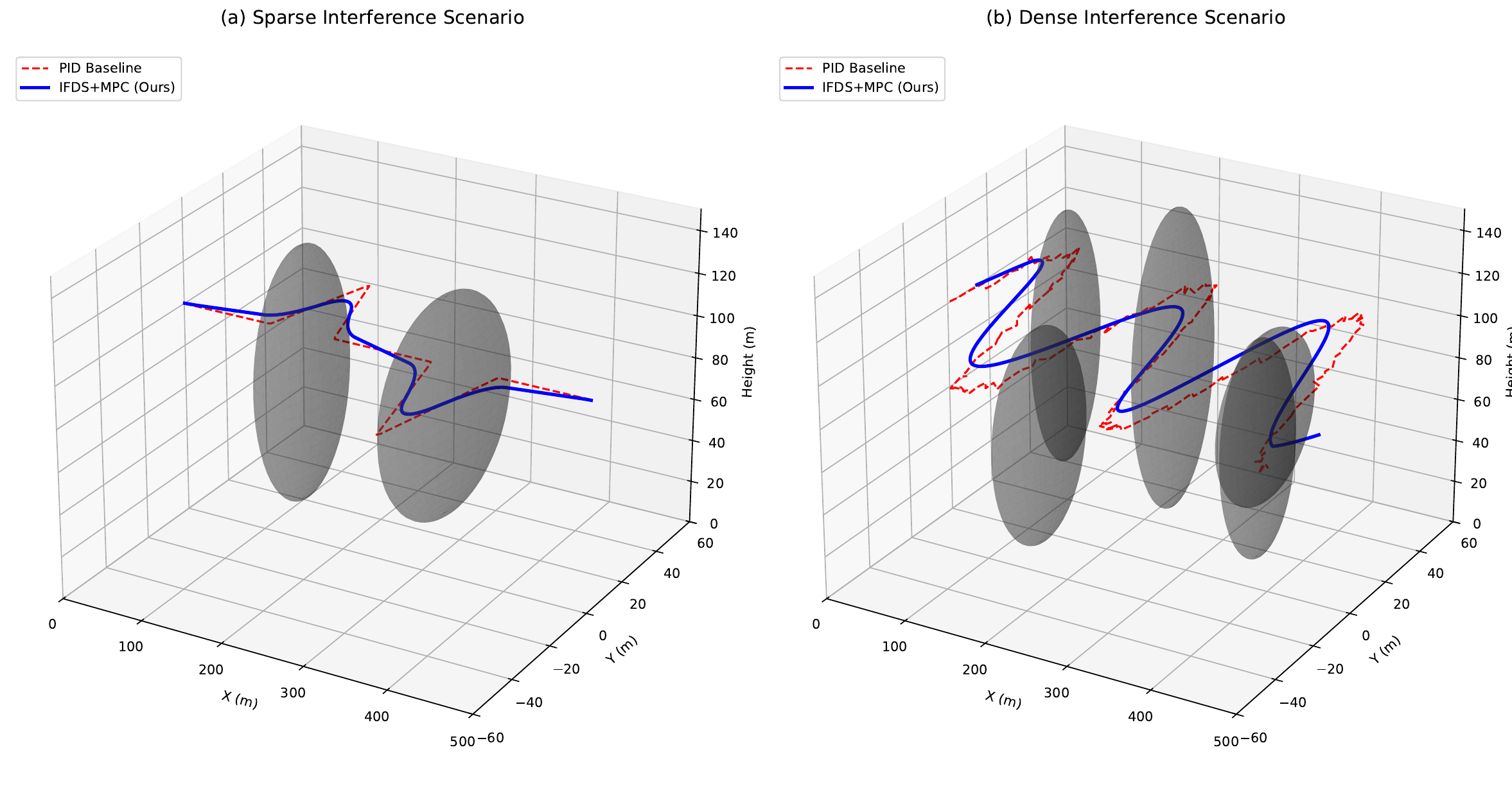}
    \caption{Comparison of UAV 3D flight trajectories in scenarios with different interference densities. Left: Sparse interference scenario; Right: Dense interference scenario.}
    \label{fig:11}
\end{figure}

Figure \ref{fig:11} shows a typical trajectory comparison in the dense interference scenario. It can be seen that the traditional PID method exhibits severe heading jitter (Overshoot) when facing sudden shadows, while the trajectory generated by the IFDS+MPC strategy has continuous curvature changes, reflecting the smooth characteristics of fluid dynamic modulation.

For quantitative evaluation, Table \ref{tab:2} summarizes the performance indicators of different algorithms, where the path smoothness indicator is defined as the integral of the rate of change of the heading angle $S = \int (\dot{\psi}^2 + \dot{\theta}^2) dt$. The results show that the IFDS+MPC method proposed in this paper improves smoothness by 42\% with only a 3.4\% increase in path length, significantly reducing the energy consumption of actuators. Meanwhile, the success rate reaches 98\%, far higher than the 68\% of the PID baseline method and 86\% of the IFDS-only method, and the minimum obstacle distance reaches 5.5m, ensuring flight safety.

\begin{table}[htbp]
    \centering
    \caption{Quantitative Comparison of Performance Indicators for Different Algorithms (Mean values in dense interference scenario)}
    \label{tab:2}
    \begin{tabular}{lcccc}
        \hline
        \textbf{Algorithm} & \textbf{Success Rate (\%)} & \textbf{Avg Path Length (m)} & \textbf{Smoothness (S)} & \textbf{Min Obs Dist (m)} \\
        \hline
        PID Baseline & 68.0 & 512.5 & 18.4 & 0.8 (Collision) \\
        IFDS Only & 86.0 & 535.2 & 12.1 & 4.2 \\
        IFDS + MPC (Ours) & 98.0 & 528.6 & \textbf{7.8} & \textbf{5.5} \\
        \hline
    \end{tabular}
\end{table}

\subsubsection{Key Parameter Sensitivity and Real-time Analysis}
\begin{figure}[htbp]
    \centering
    \includegraphics[width=0.9\linewidth]{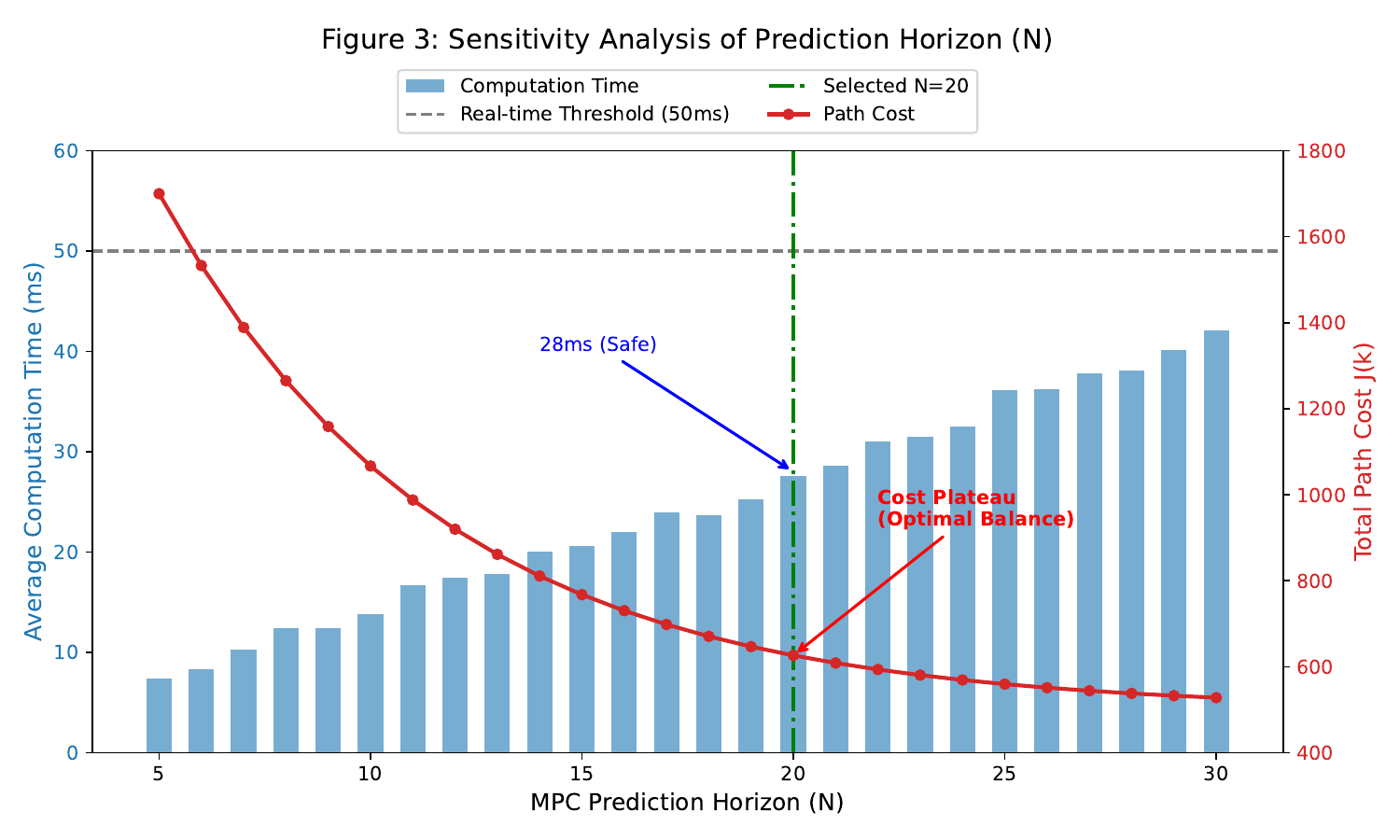}
    \caption{Impact of prediction horizon $N$ on algorithm real-time performance and path quality. The line graph represents the total path cost, and the bar chart represents the single-step computation time.}
    \label{fig:12}
\end{figure}

The MPC prediction horizon $N$ is a key parameter balancing computational load and planning optimality, directly affecting the feasibility of algorithm deployment in actual water quality monitoring tasks. Therefore, we tested the system performance within the range of $N \in [5, 30]$, and the results are shown in Figure \ref{fig:12}.

\begin{itemize}
    \item \textbf{Path Quality:} As $N$ increases, the UAV can perceive moving shadows ahead earlier, thereby making smoother avoidance maneuvers, and the total path cost shows a downward trend. When $N > 20$, the cost reduction tends to plateau.
    \item \textbf{Real-time Performance:} Computation time increases almost linearly with $N$. On the test platform (Intel i7-10700K), when $N=20$, the average single-step time is 28ms, which is far less than the system sampling period $\Delta T = 100ms$. This satisfies the real-time calculation requirements on the airborne side and ensures the continuous execution of water quality monitoring tasks.
\end{itemize}

Overall, selecting $N=20$ can obtain an observation path close to the global optimum while ensuring a control frequency of over 35Hz, achieving the optimal balance between path quality and real-time performance.

\subsection{DFAA Mechanism Ablation Simulation}
\begin{figure}[htbp]
    \centering
    \includegraphics[width=0.8\linewidth]{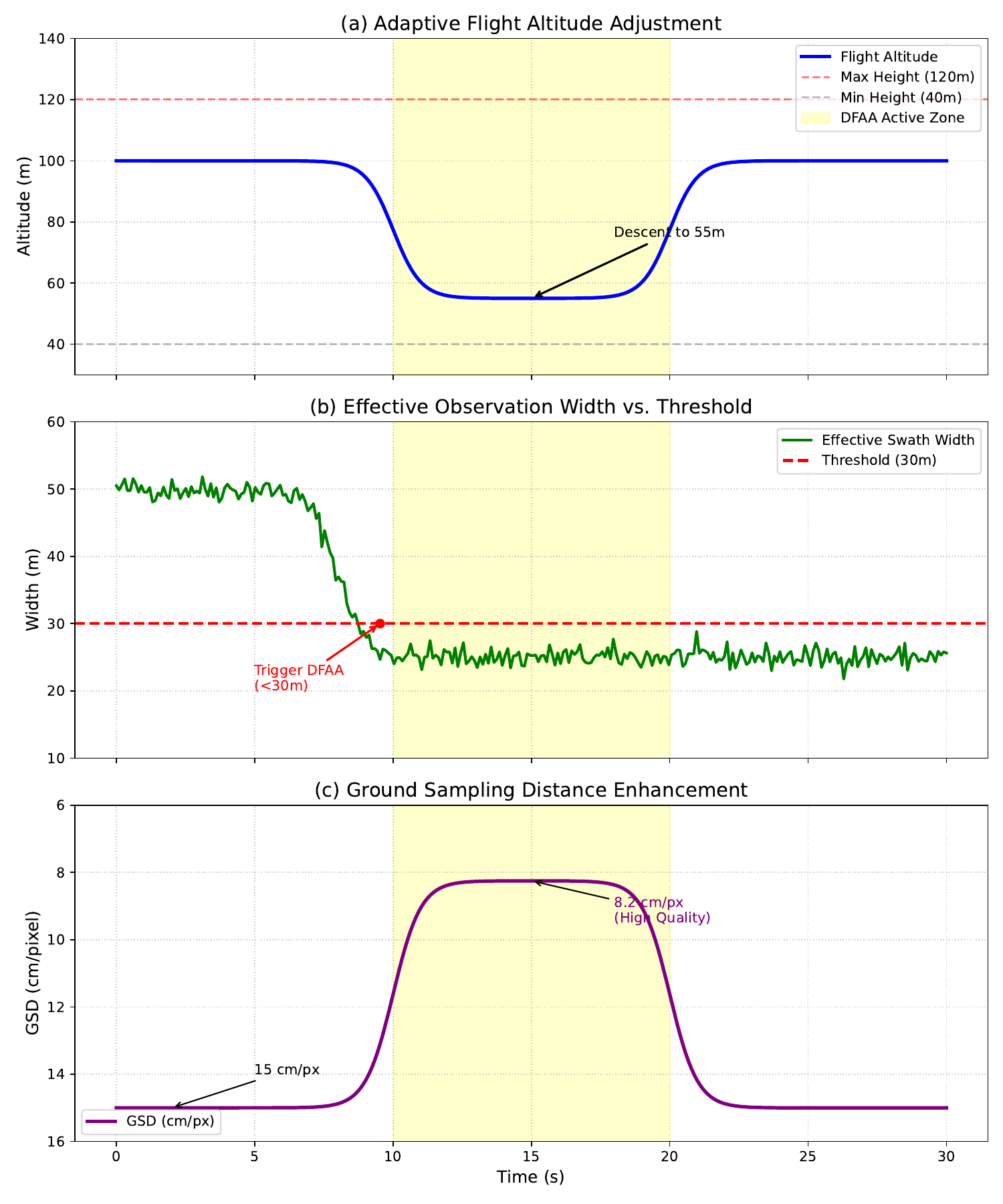}
    \caption{Analysis of the impact of DFAA mechanism on observation quality. In the interval $t = 10s - 20s$, the system detects that the effective observation width $W_{eff}$ is lower than the threshold and actively lowers the altitude to improve GSD resolution.}
    \label{fig:13}
\end{figure}

This section focuses on verifying the impact of the Dynamic Flight Altitude Adjustment (DFAA) mechanism on the quality of water quality monitoring data. Figure \ref{fig:13} shows the time-varying relationship between flight altitude, effective observation width (Swath Width), and Ground Sampling Distance (GSD) during a passage through a narrow non-shadow channel.

Simulation results indicate that when the effective observation width $W_{eff} < 30m$ is detected, the DFAA mechanism guides the UAV to descend smoothly from 100m to 55m. This action improves the Ground Sampling Distance (GSD) of the acquired spectral images from 15cm/pixel to 8.2cm/pixel, thereby enabling the capture of finer water pollution distribution characteristics. Compared with fixed-altitude flight, the effective data acquisition volume under this strategy increased by about 27\% (see previous basic experiment result analysis in Figure \ref{fig:10}), responding to the core demand for high-resolution data in water quality detection tasks and verifying the effectiveness of "quality-driven" path planning.

\subsection{Robustness Test: Influence of Environmental Perception Noise}
Considering the errors that may exist in meteorological sensors and EKF filtering in actual flight, to verify the reliability of the algorithm in real complex environments, Gaussian white noise with a mean of 0 and a standard deviation of $\sigma$ was introduced into the obstacle position observation to test the system's adaptability to perception uncertainty. Experimental results show that thanks to the flow field repulsion characteristics of IFDS, the system can still maintain a collision-free success rate of 92\% even when the position perception error reaches $\sigma = 3m$ (i.e., 20\% of the obstacle radius). This proves that modeling obstacles as super-ellipsoids with a safety margin $\gamma_{safe}$, combined with the rolling optimization strategy, can effectively offset the risks brought by perception uncertainty and ensure the stable application of the algorithm in actual water quality monitoring scenarios.

\section{Conclusion}
Addressing the common issue of light interference in UAV water quality monitoring tasks, this paper proposes a dynamic path planning method that integrates the Interfered Fluid Dynamical System (IFDS) and Model Predictive Control (MPC). By modeling time-varying shadows and reflective areas as dynamic virtual obstacles and incorporating the Dynamic Flight Altitude Adjustment (DFAA) strategy, the system achieves ``quality-driven'' intelligent flight decision-making. The main conclusions are as follows:

\begin{enumerate}
    \item The proposed hybrid control strategy effectively resolves the difficulty traditional rigid path planning faces in adapting to dynamic unstructured environments. IFDS ensures the smoothness of obstacle avoidance paths by utilizing fluid flow characteristics, while MPC solves the optimal control problem under multiple constraints within milliseconds through rolling optimization.
    \item Simulation experiments demonstrate that, while ensuring flight safety (with a 98\% success rate), the method increases the coverage area of effective water quality monitoring data by approximately 27\% through active avoidance of interference zones and adaptive altitude adjustment.
    \item Robustness analysis confirms that the algorithm maintains high reliability even in the presence of environmental perception noise.
\end{enumerate}

Future work will focus on extending the method to multi-UAV collaborative monitoring scenarios, as well as conducting hardware-in-the-loop testing and application verification in real-world complex water environments.

\bibliography{ref}
\bibliographystyle{elsarticle-harv}



\end{document}